\documentclass{article}
\usepackage{CJKutf8}
\usepackage[utf8]{inputenc}
\usepackage{multirow}
\usepackage{placeins}
\usepackage{graphicx}
\usepackage{soul}
\usepackage{xcolor}

\title{GPT-based Generation for\\ Classical Chinese Poetry \footnote{A side product by Huawei Noah's Ark Lab when  investigating the capability of Generative Pretrained Language Model (GPT).}}
\author{ Yi Liao, Yasheng Wang, Qun Liu, Xin Jiang\\
[1.5ex]%
Huawei Noah's Ark Lab} 
\date{June 2019}

\begin{document}
\begin{CJK*}{UTF8}{gbsn}
\maketitle
\begin{abstract}
    We present a simple yet effective method for generating high quality classical Chinese poetry with Generative Pre-trained Language Model (GPT)\cite{gpt}. The method adopts a simple GPT model, without using any human crafted rules or features, or designing any additional neural components. While the proposed model learns to generate various forms of classical Chinese poems, including Jueju(绝句), L\"{u}shi(律诗), various Cipai(词牌) and Couples(对联), the generated poems are of very high quality. We also propose and implement a method to fine-tune the model to generate acrostic poetry. To the best of our knowledge, this is the first to employ GPT in developing a poetry generation system. We have released an online mini demonstration program on Wechat\footnote{User may have to register a Wechat account and add ``EI体验空间" or ``诺亚实验室"} to show the generation capability of the proposed method for classical Chinese poetry.
\end{abstract}
\section{Introduction}
Classical Chinese poetry generation is an interesting challenge of natural language generation. Unlike free text generation, a classical Chinese poem should normally meet both \textit{form} and \textit{content} requirements \cite{jiuge}. The \textit{form} requirements includes the regulations on the number of words (字数), rhyming (押韵), tone patterns (平仄), pairing (对仗), etc . The other requirement is regarding \textit{content}, which requires that the theme of a poem is consistent and coherent throughout the poem.

There are many different forms of classical Chinese poetry.  Appendix A gives a brief introduction of these forms.  Our system mainly focus those forms which have strict rules, which include couplets(对联)\footnote{Couplets are not regarded as poems normally but here we do so just for convinience.}, Wujue(五绝), Qijue(七绝), Wul\"{u}(五律), Qil\"{u}(七律), and vaious Cipai(词牌) including Xijiangyue(西江月), Manjianghong(满江红), Shuidiaogetou(水调歌头), etc .

Various methods e.g., \cite{jiuge,juntao} have been proposed to generate classical Chinese poetry. However, these methods are somewhat complicated so as to satisfy the aforementioned requirements in both form and content. For example, template-based or constraint checking method is employed to guarantee the correctness of the form of the generated poetry. Key-words based mechanism is proposed to guarantee the consistency and coherency of a poem. 

In this paper, we study the problem of poetry generation given a specific type of form requirement and a specific theme. In contrast with the existing methods, we propose a poetry generation method based on the pre-trained model GPT. The underlying model of our proposed method is simply a GPT language model fine-tuned with a classical Chinese poetry corpus, without any additional modifications. All we need to do is to serialize the training poems into formatted text sequences as training data. Poems are generated by sampling from the language model token by token without any constraint to meet the form and content requirements.

In addition, we proposed a fine-tune method to train a model to generate acrostic poetry (藏头诗), where some characters in given positions are pre-defined. The proposed method can  guarantee that specific tokens can be generated by the language model in the corresponding positions.

Compared with existing methods, our propose method has below characteristics:
\begin{enumerate}
    \item Model Conciseness. The proposed method is a simple Transformer model without additional variables. However, it is powerful enough to guarantee the \textit{form} and \textit{content} requirements. Neither we use any human-defined rules or features, nor we define any specific designed neural networks rather the standard GPT. 
    \item  Well-formedness.  We surprisingly observe that although we  did not explicitly feed the model with any rules or features about classic Chinese poetry, such as the number of characters, rythming, tune patterns and coupling, the model is able to generate poems that automatically meet these rules very well for the tens of forms, even for some fairly complicated ``Cipai" like ``Shuidiaogetou" which contain around 100 characters. Actually, even for ordinary Chinese people,
    it is quite hard to master the skills to write well formed classical poems.
    \item Poetry Diversity. We employ truncated top-k sampling strategy during the generation process. Hence the generated poems are highly diverse in different runs given the same form and theme.
    \item Poetry Artistry. We observe that the model have a fair chance to generate high quality poems that express the poetry themes artistically, which is close to one written by specialized poets. Table \ref{tab:distinguish} shows four poems, among which only one was written by a Chinese poet more than one thousand years ago, while the remaining three poems are generated by our system. 
\end{enumerate}
\begin{table}[ht]
    \centering
\begin{tabular}{c}\hline
江上田家\\
村南喧鸟雀，江北梦悠悠。\\
桑熟蚕三眠，人家半依楼。\\
一身千万里，何处得穷愁。\\
日暮歌明月，长河满斛秋。
\\ \hline

江上田家\\
江边田舍好，茅屋远相迎。\\
竹里开门入，芦中引水行。\\
犬来沙上吠，鸥去岸间鸣。\\
不是无吟兴，谁知乐此生。
\\ \hline
江上田家\\
近海川原薄，人家本自稀。\\
黍苗期腊酒，霜叶是寒衣。\\
市井谁相识，渔樵夜始归。\\
不须骑马问，恐畏狎鸥飞。
\\ \hline
江上田家\\
野水通渔路，江村带夕阳。\\
数家深竹里，一树隔芦塘。\\
牧去牛羊下，人行果橘旁。\\
相逢皆贺岁，还有醉眠乡。\\ \hline
\end{tabular}
    \caption{A real poem and three poems generated by our system}
    \label{tab:distinguish}
\end{table}
\section{Our Method}
\subsection{Model Details}
\begin{figure}
    \centering
    \includegraphics[scale=0.55]{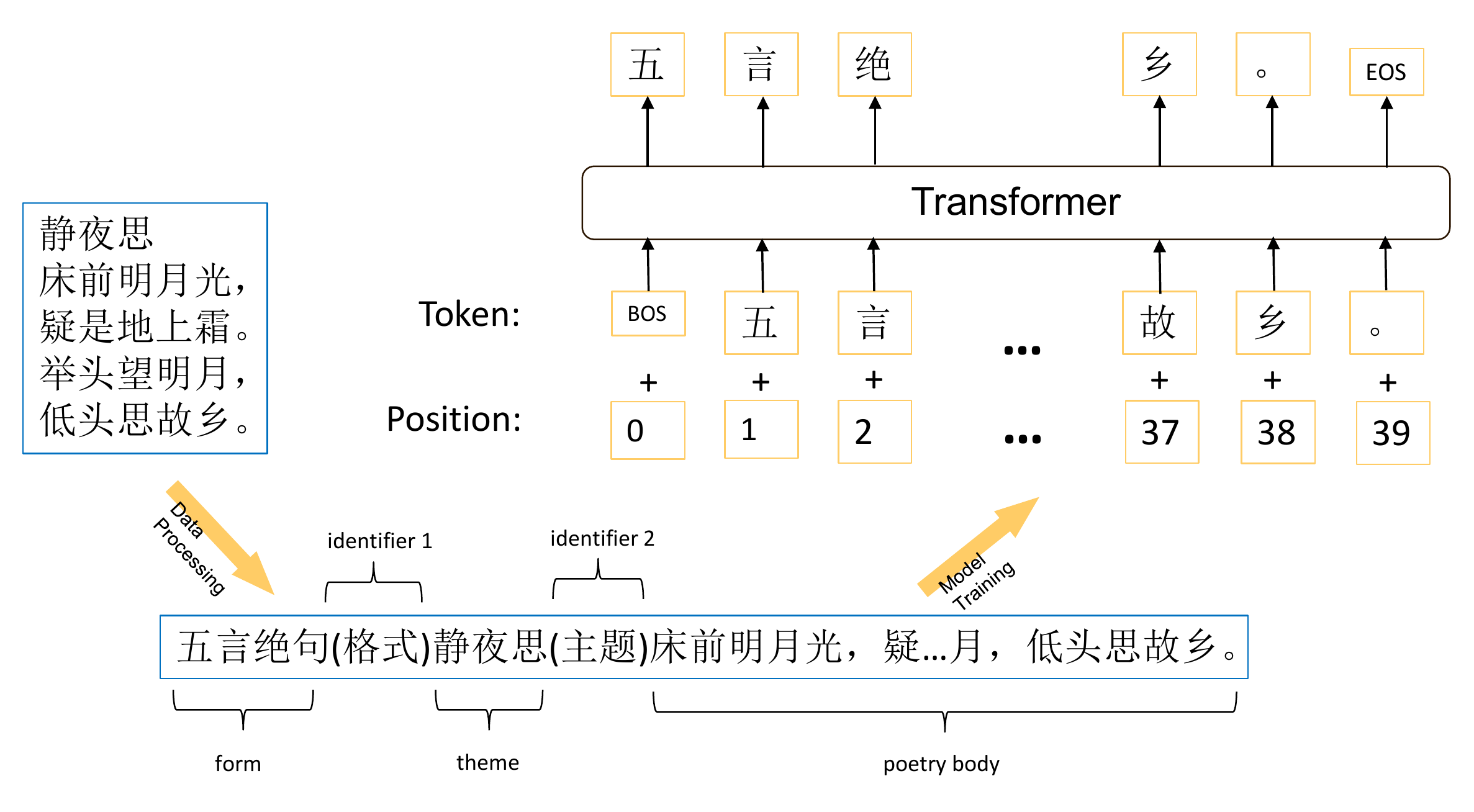}
    \caption{The process for training the poetry generation model}
    \label{fig:overview}
\end{figure}
We refer the readers to the blog \footnote{http://jalammar.github.io/illustrated-transformer/} or the papers \cite{attention, gpt, bert} to get a basic understanding of Transformer, which is the underlying model of the proposed method. Figure \ref{fig:overview} depicts the process of training the poetry generation model.

We implement our own GPT model based on the source code of BERT\footnote{https://github.com/google-research/bert}. The configuration of the size of the transformer is identical to the BERT-Base. We also adopt the tokenization script and Chinese vocab released in BERT. For text generation, we implement truncated top-k sampling instead of beam-search to generate diverse text \cite{top-k}. 

\subsection{Data processing}
The training process includes two phases: pre-training and fine-tuning. 
Our GPT model is pre-trained with a Chinese news corpus.

For fine-tuning, we collect publicly available classical Chinese poems. As shown in Figure \ref{fig:overview}, a sample poem is first transformed into a formatted sequence. The sequence contains three main fields: \textit{form}, \textit{theme}, and \textit{peotry body}. An identifier is inserted to separate two fields. Note that two different identifiers are used separate forms from themes, and themes from bodies, respectively. After preprocessing, all the training samples are text sequences in the format like [\textit{form}, \textit{identifier 1}, \textit{theme}, \textit{identifier 2}, \textit{body}].

A special case is couplets.  In most case couplets do not have a theme, we use the first line as the theme and the second line as the body.  The \textit{form} is automatically filled with ``对联" (couplets). So the generation of a couplet becomes to generate the second line given the first line, which exactly mimics the activity of Duiduizi (对对子).

The statistics of the pre-training data and fine-tuning data of our model are given in Table \ref{tab:statistics of training data}.
\begin{table}[]
    \centering
    \begin{tabular}{c|c|c}
    \hline
    Training Phases & Corpus type & Corpus size \\ \hline
        Pre-training & Chinese news& 235M sentences\\\hline
        \multirow{3}{*}{Fine-tuning} &Jueju and L\"{u}shi& 250,000 Jueju and L\"{u}shi\\ 
        &Cipai& 20,000 Cis\\ 
        &Couplet& 700,000 pairs of couplets\\ \hline
    \end{tabular}
    \caption{Statistics of Training Data and Training epochs}
    \label{tab:statistics of training data}
\end{table}

\subsection{Model Training}

\textbf{Pre-training}: We pre-trained our GPT model on Huawei Cloud Service with a news corpus which is detailed in Table \ref{tab:statistics of training data}. We trained it in 8 Nvidia V100 (16 GB) GPUs for 4 epochs. The pre-training takes totally 90 hours\footnote{The model underfitted the data when we employed it for fine-tuning}. Chinese Wikipedia can be an alternative training corpus.

\textbf{Fine-tuning}: We feed the all the training sequences of poems into the transformer and train a auto-regressive language model. The objective is to maximize the probability of observing any sequence $X = \{x_1, x_2, ..., x_{|X|}\}$:
\begin{equation}
    P(X) = \sum_{1 \leq i \leq |X|} \log p(x_i|x_1,...,x_{i-1})
\end{equation}
\noindent where $p(x_i|x_1,...,x_{i-1})$ is the probability that the token $x_i$ will be generated given all the historical tokens. The fine-tune process takes much less time as the model gets overfitted if trained too long. When the model is overfitted, it tends to retrieve raw sentences from the corpus during the generation process.

\subsection{Poetry Generation}
Once the training is completed, we apply the model to generate poems given a form requirement and a theme in the following process. We first transform the form and theme into an initial sequence as [\textit{form}, \textit{identifier 1}, \textit{theme}, \textit{identifier 2}], then the initial sequence is fed into the model and the remaining field of \textit{body} is decoded token-by-token. Note that we do not apply hard constraint during the decoding process to guarantee the correctness of the form. Instead, the model is able to automatically assign high probabilities to commas and periods in certain positions when decoding. The decoding process is end when we reach the end of the body, which is recognized by an ``EOS" token. 

\textbf{Truncated top-k sampling:} Instead of beam-search during the decoding process, we apply truncated top-k sampling strategy to obtain diverse poems. Each time to sample a token, tokens with top-k largest probabilities are first selected and then a specific token is sampled from the top-k tokens. We observe that the generated poems are in correct form even though truncated top-k sampling strategy is applied.

\subsection{Train a model for acrostic poetry generation}
We employ the same method for train a model for generating acrostic poetry. While the training and generation processes are exactly the same, we format the sequence in a slightly different way. Specifically, we replace the original theme of a poem with the combinations of the first character in each line. In the example, the first characters are ``床", ``疑", ``举", and ``低". Then the original them ``静夜思" is replaced with ``床疑举低". With the new data processing method, this training sample becomes ``  五言绝句(格式)床疑举低(藏头诗)床前明月光，疑…月，低头思故乡。"

\section{Generated examples and observations}

A selected set of generated examples by our model is given in Table \ref{tab:couplet} - \ref{tab:acrostic}. 

We observed that:
\begin{itemize}
    \item The method performs consistently well in generating  couplets(对联), jueju(绝句) and l\"{u}shi(律诗). For couplets shown in Table \ref{tab:couplet}, almost all the generated second line (下联) are paired with their corresponding first line (上联) in terms of characters in the same position. For L\"{u}shi as shown in the third and forth poetry in Table \ref{tab:shi}, it pairs the third sentences and the forth sentences, and pairs the fifth and sixth sentences, while the remaining sentences are not paired. The observation is quite surprising as the model learns the much complicated pairing rules for L\"{u}shi, which are hard to grasp even for normal educated Chinese native speakers. Pairing sentences in certain places in a poem greatly improves its the quality and beauty.
    \item Well-formedness. More than 95\% of the generated Jueju and L\"{u}shi are well-formed as the form requirement of these poetry categories are relatively simple compared with Cipai. In terms of Cipai, the method does not performs as good as Jueju and L\"{u}shi regarding well-formedness. Possible reasons may include the complexity of the forms and the lack of sufficient data for each type of Cipai.  There are tens of thousands of training samples for each category of Jueju/L\"{u}shi, while for Cipai, there are totally 882 types of Cipai in the training corpus but only 104 of them have more than one hundred training samples each and the largest type contains only 816 training samples. The possibilities to generate correct Cis also vary for different types of Cipai, which could be attributed to the differences in complexity as well as the numbers of training samples of that type in the training corpus. The Cipai Shuidiaogetou, which contains 744 training samples, is relatively difficult as the length requirement for each line varies. Roughly 70\% of the generated Cis for Shuidiaogetou are correct in form. One of the example is shown in Table \ref{tab:ci}.
    \item Although not explicitly modelled, the rhyming (平仄) and tone patterns (押韵) of the generated poems are also fairly good. 
    \item Diversity. As we adopt sampling strategy while decoding, our method can generate highly diverse output in different runs. As shown in Table \ref{tab:couplet} and Table \ref{tab:shi}, the model generate totally different second lines (下联) for the same first line (上联) and different poems for the same theme in multiple runs. We also notice that when the given first line is in the training corpus, it is possible that the model retrieves the whole original second line from the training corpus. However, for poetry generation, it generates new sentences even though the given theme is in the training corpus.
    \item Artistry. We observe that the model can sometimes generate high quality poems that express the poetry themes artistically. However, while the generation quality for some given themes are constantly good in multiple runs, some themes, such as ``机器翻译", which appear rarely in the training corpus are less likely to generate good poems. ``秋思" is a good theme to generate high quality poems. The examples in Table \ref{tab:shi} are generated in one run without manual selection.
\end{itemize}

\begin{table}[ht]
    \centering
\begin{tabular}{c|c}
\hline 
上联 (First line) & 下联 (Second line)\\ \hline
\multirow{3}{*}{一句相思吟岁月 } & 几分寂寞醉诗词 \\
&三杯落泪诉年华 \\
&三分眷恋到天涯 \\ \hline
\multirow{3}{*}{风弦未拨心先乱 } & 诗卷未题笔已枯 \\
&月盏虽清梦已酣 \\
&草色初青眼欲穿 \\ \hline
\multirow{3}{*}{海上飞燕飞上海} & 城外环山环外城 \\
&江中落叶落中江 \\
&山前雾气雾前山 \\ \hline
\multirow{3}{*}{水墨丹青，烟雨江南春纵笔}&花红柳绿，桃源粤地燕裁云\\
&诗词歌赋，风花塞外夏抒怀\\
&天光风韵，云霞岭上日倾杯\\ \hline

\end{tabular}
    \caption{Example of Couplets. We present three generated second lines (下联) for each given first line (上联)}
    \label{tab:couplet}
\end{table}

\begin{table}[ht]
    \centering
\begin{tabular}{c}
\hline
五绝(Wujue)·秋思\\
\hline
暮燕翻惊户，\\
飞鸿却唤人。\\
西风卷梧叶，\\
触落一庭秋。\\ \hline

七绝(Qijue)·秋思\\
\hline
年华冉冉飞无翼，\\
风物萧萧滞故乡。\\
万里重云正愁绝，\\
洞庭湖外见清霜。\\ \hline

五律(Wul\"{u})·秋思 \\
\hline
秋窗堕清影，默坐不复收。\\
风来户外闹，月过竹边幽。\\
得失已搔首，是非空置忧。\\
南鸿若忘机，直飞念其幽。\\ \hline

七律(Qil\"{u})·秋思\\
\hline
南浦西山雨送秋，病留白发日添愁。\\
风尘苒苒侵儒素，霜露番番到客裘。\\
客子渐衰空感慨，古人难得始淹留。\\
欲谈往事浑无可，万里波涛一叶舟。\\ \hline
\end{tabular}
    \caption{Theme ``秋思" with different form requirements. Generated in one run without manual selection}
    \label{tab:shi}
\end{table}

\begin{table}[ht]
    \centering
\begin{tabular}{l}
\hline
\multicolumn{1}{c}{水调歌头 (Shuidiaogetou)·游园} \\ 
\hline
天上摘星汉，谪下笔端严。\\
云台望断，千古遗恨沓难兼。\\
谁信而今老态，犹恋吾庐清磐，乘兴下江南。\\
便好徜徉去，好处却容瞻。\\
\\
风流事，人未老，气方炎。\\
平生出处何似，此乐几年淹。\\
闻道玉霄仙驾，犹记横空丹鹤，飘下五云顶。\\
不是出世相，来到玉皇边。\\
\hline
\end{tabular}
    \caption{Examples of Ci。}
    \label{tab:ci}
\end{table}

\begin{table}[ht]
    \centering
\begin{tabular}{c}\hline
七律(Qil\"{u})·一路平安\\
\hline
一声天际雁横秋，忽梦青城旧友游。\\
路入青林无去马，手携黄牒有归舟。\\
平生志业商山老，何日公卿汉署留。\\
安得相从话畴昔，一樽同醉万山头。\\
\hline
\end{tabular}
    \caption{Examples for Acrostic Poetry}
    \label{tab:acrostic}
\end{table}

\section{Related Work}
Early works \cite{yan2013, zhou2012, wu2019, zhou2010} on Chinese poetry generation have been mostly rule-based or template-based.  Recurrent Neural
Network (RNN) \cite{rnn1} was recently introduced as it has been proved to be effective in generation tasks such as machine translation and dialog generation. However, few researchers have adopted the latest self-attention models for  generating poems. As far as we know, we are the first to employ GPT in developing a poetry generation system. GPT has been famous for having the capability to generate text that can hardly be distinguished even by human beings. As a natural consequence, a GPT-based method could potentially write high quality poems. 

Existing methods have been focusing on improve the well-formedness and content coherence of generated poems. To make sure the rhyming, the tone patterns and the pairing of a generation are correct, various strategies have been adopted. For example, Yan (2016) \cite{polish} proposes an iterative polishing schema, which refines the generated poem until a well-formed one is obtained. In the mean time, some other works have been investigating the coherence of content throughout a poem. For example, Yi et al. (2018) \cite{jiuge} propose a salient-clue mechanism which automatically selects the most salient characters from the so-far generated lines as a theme clue for generating the next line. Yang et al. (2017) \cite{vae} and Wang et al. (2016) \cite{plan} employ a two-stage approach, where a set of keywords are planned first and are then fed into the generation of different lines sequentially. Besides above works, some researchers investigate other interesting topics on poetry generation. For example, Yang et al. (2018) \cite{stylish} propose a model for stylistic Chinese poetry generation .

Compared with the existing methods, the major advantage of our proposed approach is the conciseness and simplicity of the model. Meanwhile, it still exhibits strong, or sometimes even better ability in generating well-formed and coherent poems. For example, it is easy for our proposed method to generate well-paired sentences at once, especially for L\"{u}shi, which, however, is relatively difficult for the existing method unless multiple times of polishing are adopted. Regarding the content coherence, in rare cases, our proposed method is even beyond relying on keywords to make the poetry coherent. Rather, it expresses either the story, the scene, or the emotion to describe a deeply coherent poetry comprehensively and artistically.

\section{Conclusions and Future Works}
We present a classical Chinese poetry generation method based on pre-trained language model. The proposed method is far simpler than existing method based on recurrent neural networks and can generate better poems in some perspectives. Though the generated poems are not perfect all the time, our preliminary experiments have shown that GPT provides a good start to promote the overall quality of generated poems. That is, how to express the scene, the story, the emotions, and so on in a natural and artistry way. We present this report in the hope of helping researchers in understanding the capability of GPT as well as developing better poetry generation systems.

\paragraph{Acknowledgements}
The authors would like to thank Yuanzhen Liu and Xiaoya Wei for building the demonstration system on Wechat; to thank Wenyong Huang, Xiaozhe Ren, and Weiguo Li for building the training platform on Huawei Cloud Service; to thank Junqiu Wei, Xiaoguang Li, Liangyou Li, Yun Chen, Meng Zhang, Yinpeng Guo and Xiao Chen for providing insightful suggestions.

\FloatBarrier

\appendix

\section{Forms of Classical Chinese Poetry}
According to the form requirements, there are different categories of classical Chinese poetry which could be summarized as follows:
\begin{itemize}
\item Couplets (对联): A couplet is a pair of sentences in classical Chinese poems which follow strict rules on lengths, rhyming, tone patterns and paring.  Couplets are normally not regarded as poems, however they can also be used independently.  Here we treat couplets as a category of poetry just for convinience. 
\item Old Style Poetry (Gutishi,古体诗): mainly includes two forms: 
\begin{itemize}
    \item Five-character Gushi (五言古诗), contain variable number of sentence pairs where each sentence has 5 characters.
    \item Seven-character Gushi (七言古诗), contains variable number of sentence pairs where each sentence has 7 characters.
\end{itemize}  Its forms are relatively flexible without strict regulations on lengths, rhyming, tone patterns and pairing.
\item New Stype Poetry (Jintishi,近体诗): mainly includes four forms: 
\begin{itemize}
    \item Five-character Jueju (or Wujue, 五言绝句，五绝)，contains 4 sentences where each sentence has 5 characters, totally 20 characters.
    \item Seven-character Jueju (or Qijue, 七言绝句，七绝), contains 4 sentences where each sentence has 7 characters, totally 28 characters.
    \item Five-character L\"{u}shi (or Qil\"{u}, 五言律诗, 五律), contains 8 sentences where each sentence has 5 characters, totally 40 characters. 
    \item Seven-character L\"{u}shi (or Qil\"{u}, 七言律诗, 七律), contains 8 sentences where each sentence has 7 characters, totally 56 characters.
\end{itemize}
There are strict rules on these forms regarding lengths, rhyming, tone patterns and paring.
\item Lyric Poetry (Ci,词): Unlike Gutishi or Jintishi, the sentences in a piece of Ci have different number of characters. Depending on the combinations of different length sentences, there are about 100 different forms of Ci (Cipai), where some of the most frequently used forms are:
\begin{itemize}
    \item Shuidiaogetou (水调歌头)
    \item Manjianghong (满江红)
    \item Langtaosha (浪淘沙)
    \item Xijiangyue (西江月)
    \item ...
\end{itemize}
Ci also has very strict rules on lengths, rhyming, tone patterns and paring.
\end{itemize}

\section{More Examples}

\begin{table}[ht]
    \centering
\begin{tabular}{c|c}
\multicolumn{2}{c}{对联(Couplets) }\\ \hline
上联 (First line)&下联 (Second Line) \\ \hline
月色湖光鱼戏月& 兰馨梅馥鹿鸣兰\\ \hline
一亭风物一亭荷& 千载人文千载歌 \\ \hline
人轻担重轻担重 & 水远山高远山高 \\ \hline
客家小镇，名家故里，中和风土毓神奇 & 广府古城，百府新店，生态园林荟锦绣 \\ \hline
美园美焕古今情，喜远养禅风，近养仁风，&山寺山幽林壑秀，欣高吟福地，低吟寿地，\\ 
远近无双朝锦绣 & 高低不尽醉芬芳 \\ \hline
\end{tabular}
    \caption{More examples of Couplets.}
    \label{tab:my_label}
\end{table}

\begin{table}[ht]
    \centering
\begin{tabular}{c}
\hline
五绝(Wujue)·空谷幽兰\\
\hline
幽谷暖生春，\\
寒花带晚雪。\\
不是香风清，\\
何人解赏节。\\ \hline

七绝(Qijue)·空谷幽兰\\
\hline
一树寒英独自芳，\\
几年寂寞在空谷。\\
清晨携手远来看，\\
惟有幽兰开不得。\\ \hline

五律(Wul\"{u})·空谷幽兰 \\
\hline
一从幽谷植，芳气袭人来。\\
为见春风好，因闻夜月哀。\\
清香飘静畹，寒蝶绕闲栽。\\
未得同兰菊，空阶自往回。\\ \hline

七律(Qil\"{u})·空谷幽兰\\
\hline
一声清梵彻中林，暗吐芳兰气欲深。\\
幽谷乍闻风淅沥，空门应得露沈沈。\\
香花散乱飘残雪，尘迹荒凉断去音。\\
我有孤根在何处，春山遥夜月初沉。\\ \hline
\end{tabular}
    \caption{Theme ``空谷幽兰" with different form requirements.}
\end{table}

\begin{table}[ht]
    \centering
\begin{tabular}{c}
\hline
五绝(Wujue)·瑞雪兆丰年\\
\hline
瑞雪中山酒，\\
农祥吉岁功。\\
从今三白兆，\\
大府似新丰。\\ \hline

七绝(Qijue)·瑞雪兆丰年\\
\hline
十分晴色弄春寒，\\
雪压云低晓更看。\\
农事未明天意好，\\
一蓑带月下长滩。\\ \hline

五律(Wul\"{u})·瑞雪兆丰年 \\
\hline
雪兆开丰岁，农祥积庆家。\\
三登禾稼熟，万顷玉粒加。\\
晓色晴方好，风光日欲斜。\\
愿流鸿满野，飞洒到天涯。\\ \hline

七律(Qil\"{u})·瑞雪兆丰年\\
\hline
雪后乾坤万象开，老天还有喜春回。\\
三农已报丰年瑞，百兽应知温室梅。\\
千里欢声连绿野，几家高卧卷青苔。\\
麦田多庆须吾辈，闻说江东亦好栽。\\ \hline
\end{tabular}
    \caption{Theme ``瑞雪兆丰年" with different form requirements.}
\end{table}

\begin{table}[ht]
    \centering
\begin{tabular}{c}
\hline
五绝(Wujue)·溪流背坡村\\
\hline
溪流向人急，\\
山路背城斜。\\
不识春林处，\\
蜂喧蝶趁衙。\\ \hline

七绝(Qijue)·溪流背坡村\\
\hline
一湾窈窕竹婵娟，\\
十里平波镜样天。\\
白鹭飞来无个事，\\
淡烟和雨暝苍然。\\ \hline

五律(Wul\"{u})·溪流背坡村 \\
\hline
溪流疑有路，日暮缓跻攀。\\
白石前村近，青山远寺间。\\
雨余林翠湿，风软鸟声闲。\\
幽趣无人会，扁舟又欲还。\\ \hline

七律(Qil\"{u})·溪流背坡村\\
\hline
十里荒坡路不通，一溪流过碧玲珑。\\
树头鸦舅遥相应，篱下鸡豚晚自丰。\\
岁旱得粮多乐业，村深无酒易浇胸。\\
老翁已作扶犁叟，儿女逢迎笑语同。\\ \hline
\end{tabular}
    \caption{Theme ``溪流背坡村" with different form requirements.}
\end{table}

\begin{table}[ht]
    \centering
\begin{tabular}{c}
\hline
五绝(Wujue)·沧海横流方显英雄本色\\
\hline
沧海横丈夫，\\
慷慨平生气。\\
胸中一寸铁，\\
霹雳惊鱼艇。\\ \hline

七绝(Qijue)·沧海横流方显英雄本色\\
\hline
一叶飘然万里秋，\\
独凭江槛思悠悠。\\
英雄只有心相许，\\
到处须横反不流。\\ \hline

五律(Wul\"{u})·沧海横流方显英雄本色 \\
\hline
赤子观星陨，沧波起风涛。\\
怒号天失夜，冻缆海变潮。\\
神力与柔颉，雄心生鼓刀。\\
一身千亿骨，敢以不羁高。\\ \hline

七律(Qil\"{u})·沧海横流方显英雄本色\\
\hline
一线春蚕起两眠，赤鳞红尾血流年。\\
长江风浪惊天地，老海波涛入圣贤。\\
此日英英呈伟画，他时咄咄逼人肩。\\
凭谁为唤真头角，直上冲云跨九天。\\ \hline
\end{tabular}
    \caption{Theme ``沧海横流方显英雄本色" with different form requirements.}
\end{table}

\begin{table}[ht]
    \centering
\begin{tabular}{c}
\hline
五绝(Wujue)·端午\\
\hline
节遇端阳日，\\
蒲觞满自倾。\\
儿童采菖屑，\\
白首忆家庭。\\ \hline

七绝(Qijue)·端午\\
\hline
榴花角黍泛菖蒲，\\
想见端居昼景舒。\\
欲赋前贤无好语，\\
愿将此意答天衢。\\ \hline

五律(Wul\"{u})·端午 \\
\hline
节序重重过，乡邻草草同。\\
不堪时服彩，空把角巾东。\\
江月随人好，荆花笑己穷。\\
今朝千里至，共贺两年丰。\\ \hline

七律(Qil\"{u})·端午\\
\hline
彩丝百缕纫为佩，艾叶千窠结作人。\\
台架周遭珠竞出，宫庭左右玉争新。\\
香浓蜜渍蜂房细，味硬金蒸蚁杓醇。\\
欲学灵均吊湘客，不知谁是楚骚臣。\\ \hline
\end{tabular}
    \caption{Theme ``端午" with different form requirements.}
\end{table}

\begin{table}[ht]
    \centering
\begin{tabular}{c}
\hline
五绝(Wujue)·春草\\
\hline
日暖春草长，\\
游子念归程。\\
此日草堂下，\\
思君肠断声。\\ \hline

七绝(Qijue)·春草\\
\hline
绿到江干草又青，\\
可怜春色太分明。\\
无情最是黄莺语，\\
犹恋斜阳犹恋名。\\ \hline

五律(Wul\"{u})·春草 \\
\hline
草色连朝尽，花枝一夜稀。\\
春光已无限，客行何必归。\\
野兴寻溪宿，乡情在鸟飞。\\
山中有羁客，应识异乡衣。\\ \hline

七律(Qil\"{u})·春草\\
\hline
几年曾记翠离宫，草木萋萋色更浓。\\
画出晓霞明涧户，笔成春雨落池峰。\\
池边好鸟冲花出，亭下闲人卷箔从。\\
闻道斜阳已惆怅，不堪空听子规钟。\\ \hline
\end{tabular}
    \caption{Theme ``春草" with different form requirements.}
\end{table}

\begin{table}[ht]
    \centering
\begin{tabular}{c}
七绝(Qijue)·机器翻译\\
\hline
机器中藏千里波，\\
无风自动亦长讹。\\
不知谁是翻译手，\\
妙用方圆宛转多。\\ \hline
\hline
\end{tabular}
    \caption{More examples of Qijue}
    \label{tab:my_label}
\end{table}

\begin{table}[ht]
    \centering
\begin{tabular}{c}
七绝(Qijue)·阅读理解\\
\hline
理随心解妙难知，\\
识尽偏圆始析离。\\
若谓二三皆正用，\\
不应礼义总成违。\\
\hline
\end{tabular}
    \caption{More examples of Qijue}
    \label{tab:my_label}
\end{table}

\begin{table}[ht]
    \centering
\begin{tabular}{c}
五绝(Wujue)·华为雄起\\
\hline
\textbf{华}月移樽侧\\
\textbf{为}农古汉家\\
\textbf{雄}风生两腋\\
\textbf{起}舞纵天沙\\
\hline
Note: 两腋风生 is an idiom.\\
\hline
\\
七律(Qil\"{u})·小城故事 \\ \hline
\textbf{小}园东第起征轩，水柳低垂几万千。\\
\textbf{城}里东西新宰相，楼中左右旧王孙。\\
\textbf{故}人不见乌衣巷，晚节谁如白乐园。\\
\textbf{事}去节残人自老，花残叶落独魂销。\\ \hline
\end{tabular}
    \caption{More examples for Acrostic Poetry}
    \label{tab:my_label}
\end{table}

\begin{table}[ht]
    \centering
\begin{tabular}{c}\hline
满江红(Manjianghong)·游园\\ \hline
一雨连朝，春事少、游园病酒。\\
还记取、少年游乐，踏青斗草。\\
红日三竿花似洗，玉肌十二香如扫。\\
任醉扶、金盏倒垂莲，都忘了。\\
\\
狂态度，无拘检。\\
娇姿媚，浑如削。\\
任宝钗零落，翠鬟偷帖。\\
帘外卖花枝上燕，尊前醉玉枝头小。\\
问此花、不肯转诗肠，须频绕。\\\hline
\\
\hline
武陵春(词牌名)游园(标题)\\ \hline
长忆西湖湖上宴，一笑倒琼彝。\\
红翠照人如有思，相见又相疑。\\
别岸潮生苹末涨，催发浪花飞。\\
照酒红妆间舞衣，天远水长迟。\\ \hline
\end{tabular}
    \caption{More examples for Ci}
    \label{tab:my_label}
\end{table}

\end{CJK*}
\end{document}